\documentclass[letterpaper, 10 pt, conference]{ieeeconf}  

\IEEEoverridecommandlockouts                              

\usepackage{amsmath} 
\usepackage{amssymb}  

\usepackage[algoruled]{algorithm2e}
\usepackage{amsmath}
\usepackage{amssymb}
\usepackage{caption}
\usepackage{footmisc}
\usepackage{graphicx}
\usepackage{listings}
\usepackage{subcaption}
\usepackage{xcolor}

\definecolor{backcolor}{rgb}{0.95,0.95,0.92}
\definecolor{codegreen}{rgb}{0,0.6,0}
\lstdefinestyle{mystyle}{
    commentstyle=\color{codegreen},
    keywordstyle=\color{magenta},
    stringstyle=\color{codepurple},
    basicstyle=\ttfamily\footnotesize,
    breakatwhitespace=false,         
    breaklines=true,                 
    captionpos=b,                    
    keepspaces=true,                 
    numbers=left,                    
    numbersep=5pt,                  
    showspaces=false,                
    showstringspaces=false,
    showtabs=false,                  
    tabsize=2
}
\title{\LARGE \bf Deep Interpretable Models of Theory of Mind
}%

\author{Ini Oguntola, Dana Hughes, and Katia Sycara
\thanks{* All authors are with the School of Computer Science, Carnegie Mellon University, Pittsburgh, PA, USA {\tt\small (ioguntol, danahugh, katia)@cs.cmu.edu}}%
\thanks{* This work is supported by the Defense Advanced Research Projects Agency (DARPA) under Contract No. HR001120C0036, and by the AFRL/AFOSR award FA9550-18-1-0251.}%
}

\begin{document}

\maketitle
\thispagestyle{empty}
\pagestyle{empty}

\begin{abstract}
When developing AI systems that interact with humans, it is essential to design both a system that can understand humans, and a system that humans can understand. Most deep network based agent-modeling approaches are 1) not interpretable and 2) only model external behavior, ignoring internal mental states, which potentially limits their capability for assistance, interventions, discovering false beliefs, etc. To this end, we develop an interpretable modular neural framework for modeling the intentions of other observed entities. We demonstrate the efficacy of our approach with experiments on data from human participants on a search and rescue task in Minecraft, and show that incorporating interpretability can significantly increase predictive performance under the right conditions.
\end{abstract}

\section{INTRODUCTION}

Human intelligence is remarkable not just for the way it allows us to navigate environments individually, but also how it operates socially, engaging with other intelligent actors. Humans naturally build high-level models of those around them, and are able to make inferences about their beliefs, desires and intentions (BDI) \cite{bdi}. These inferences allow people to anticipate the behaviors of others, use these predictions to condition their own behavior, and then anticipate potential responses.
In both psychology and machine learning this is referred to as \textit{theory of mind} (ToM) \cite{baker2011bayesian, rabinowitz2018machine, cuzzolin2020knowing}, which aims to model not only the external behavior of other entities but their internal mental states as well. The developmental psychology literature has found that children as young as 4 years old have already developed a ToM, a crucial ability in human social interaction~\cite{astington2010development}. ToM can enable discovery of false or incomplete beliefs and knowledge and can thus facilitate interventions to correct false beliefs. 
Therefore, work in enabling agents to develop ToM is a crucial step not only in developing more effective multi-agent AI systems but also for developing  AI systems that interact with humans, both cooperatively and competitively \cite{cuzzolin2020knowing}.

With a few exceptions~\cite{choudhury2019utility}, most agent-modeling approaches in the reinforcement learning and imitation learning literature largely ignore these internal mental states, usually only focusing on reproducing the external behavior \cite{foerster18awareness, pr2}. This limits their ability to reason in a deeper way about entities that they interact with. While prior work has explored providing agents with models of some aspect of a human's mental state, such as reward~\cite{choudhury2019utility} or rationality~\cite{shah2019feasibility}, more complex models incorporating multiple properties of mental state (e.g., beliefs over the environment, desires, personality characteristics, etc.) in non-toy environments with remains largely unexplored.

While modeling ToM of a human is a very challenging task for an artificial AI agent, understanding the reasoning of such an agent is even more challenging. This paper focuses on developing human-interpretable ToM models. We believe that the ability of a ToM model to both infer human mental states and allow humans to understand what features contribute to the model's reasoning would  
1) enable trust with humans it interacts with 2) better enables the agent to choose and \textit{justify} interventions (e.g. advice). 

In this paper we focus on modeling the intentions of observed human within a ToM framework, and discuss the potential to leverage this reasoning for \textit{interventions}, i.e. give advice to the human in the context of human-agent teaming.
We perform experiments with human trajectories obtained from simulated search and rescue tasks in a Minecraft environment, and find that enforcing interpretability can also increase predictive accuracy under the right conditions.

The primary contributions of this paper are the following:
\begin{itemize}
    \item We design a modular framework to enable AI agents to have a theory of mind model of a human.\footnote{Our method is general and supports both humans and artificial agents as observed entities. However, since our experimental evaluation is on human data, we only refer to a \textit{human} as the entity being observed.\label{footnote:humanagent}}
    \item We present a method of combining neural and non-differentiable components within our framework.
    \item We extend this framework for \textit{interpretable} intent inference with a novel application of concept whitening~\cite{conceptwhitening}.
    \item We present experimental results with human participants and provide both qualitative and quantitative interpretability analyses.
\end{itemize}

\section{RELATED WORK}

\subsection{Theory of Mind}
Theory of mind approaches to agent-modeling have shown to make inferences that align with those of human observers \cite{baker2011bayesian, scassellati2002theory}. An early work from Breazeal et. al took an approach based on embodied cognition with rule-based inference using STRIPS-like approaches from the planning literature, and were able to demonstrate false belief detection \cite{breazeal2009embodied}. Other similar early approaches also infer human goals and beliefs in the context of planning with pre-programmed non-scalable approaches \cite{oh2014probabilistic, oh2013prognostic}. Bayesian approaches to ToM \cite{baker2011bayesian} are also difficult to scale to larger environments. Rabinowitz et al. used neural networks to learn latent representations of ``mental states", but only for artificial agents in small toy environments \cite{rabinowitz2018machine}. Other work concurrent to this paper has explored theory of mind modeling of human state and behavior with graph-theoretic approaches for scenarios such as autonomous driving  \cite{chandra2020stylepredict}.

This paper presents a neural ToM approach that supports reasoning about humans, and provides experimental results within a complex environment modeling actual human decision making. In addition, we explicitly incorporate interpretability into our approach while still maintaining the performance and scalability benefits of neural approaches.

\subsection{Imitation Learning}
Theory of mind is closely related to imitation learning and inverse reinforcement learning in that both attempt to model other agents, and both can be applied to model either human or artificial agents \cite{irltom}. Although most imitation learning methods do not consider interpretability, there are approaches such as InfoGAIL that take a heuristic approach towards interpretability by using information theoretic losses to enforce structure on the latent space and discover modes in observed behavior \cite{infogail}. Others use human-readable programmatic polices for agent-modeling rather than black-box deep networks~\cite{verma2018programmatically}.
The approach presented in this paper falls under the umbrella of behavioral cloning \cite{bain1995framework, torabi2018behavioral}, with additional constraints for modeling internal states via theory of mind and for interpretable learning.
 
\subsection{Deep Interpretability}

Saliency-based methods \cite{lime, shap} are the most popular means of providing post-hoc explanations, and aim to highlight the most important input features by assigning importance weights. However, many common saliency methods have been found to be independent of both the model and the data, and often have similar resulting explanations for all inputs, which is undesirable \cite{adebayo2018sanity, rudin2019stop}. While saliency maps have also been applied in reinforcement learning contexts \cite{iyer2018transparency, annasamy2019towards}, operating on low-level input features (e.g. raw pixels) may not coincide with human ideas of explanation.

Concept-vector post-hoc methods focus on explanations based on higher-level concepts rather than low-level input features \cite{tcav, zhou2018interpretable}, but these also have their own drawbacks. In general, most post-hoc explanations are based on assumptions of the latent space that may not hold; for instance, the implicit assumptions that 1) concepts are (linearly) separable in the latent space, and that 2) each axis in the latent space only represents a single concept.

Alternatively, others have devised approaches such as concept whitening that focus on learning models that are interpretable by design \cite{rudin2019stop, koh2020concept, conceptwhitening}, i.e. approaches that shape the latent space during training. These methods have been used in classification tasks, e.g. in image classification for aligning the latent space with predefined human-interpretable concepts. In contrast to these approaches,  
this paper develops a variation of concept whitening \cite{conceptwhitening}
for modeling decision-making processes (e.g. theory of mind, imitation learning, reinforcement learning).


\section{NEURAL ToM FRAMEWORK}

\subsection{Purpose}
It is important to note that the ToM framework is used by an observer to infer mental states of an observed entity, but not to define the entity's observable behavior. The action predictions produced by the ToM model can be treated as a policy to forecast the entity's future behavior, and imitation learning can provide a training signal for the ToM model, but our purpose is \textit{not} to train an agent to perform a task.

\subsection{Overview}

\begin{figure}
    \centering
    \includegraphics[width=0.55\linewidth]{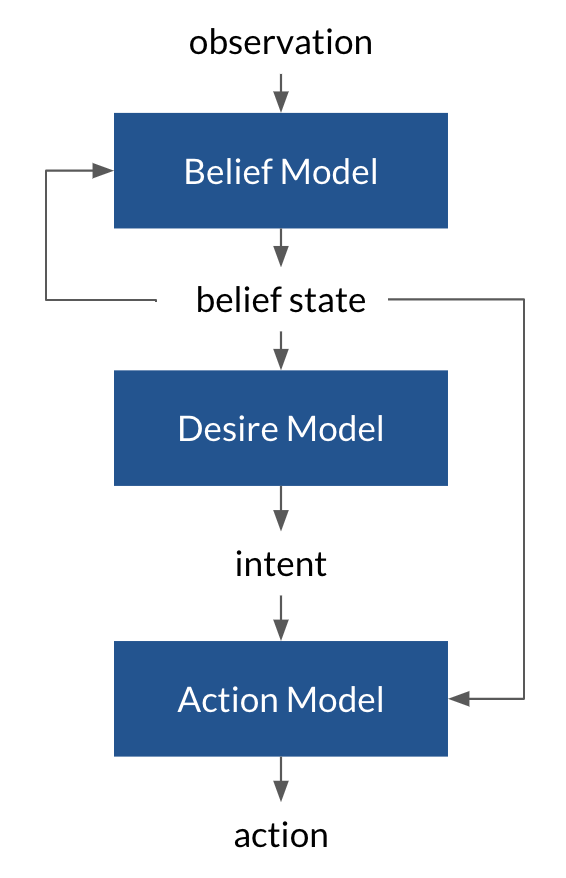}
    \caption{Modular theory of mind (ToM) framework.}
    \label{fig:modular_tom}
\end{figure}

We developed a modular theory of mind framework for an observing agent to use to infer the mental state of an observed human. As shown in Figure \ref{fig:modular_tom}, we model the decision making process of an observed entity as follows:

\paragraph{Belief Model}
The belief model $m: \mathcal{B} \times \mathcal{O} \rightarrow \mathcal{B}$ is responsible for updating the belief state. Given current belief $b \in \mathcal{B}$ and observation $o \in \mathcal{O}$, the belief model $m$ outputs an updated belief state $b_{new} = m(b, o)$.

\paragraph{Desire Model}
The desire model $g: \mathcal{B} \rightarrow \mathcal{I}$ is responsible for calculating the intent $z \in \mathcal{I}$ given updated belief $b$; that is, $g(b) = z$.

\paragraph{Action Model}
The action model $f: \mathcal{B} \times \mathcal{I} \rightarrow \mathcal{A}$ is responsible for generating or predicting an action $a \in A$, given the belief state $b$ and intent $z$; that is, $f(b, z) = a$.

The modules in our framework impose an inductive bias that reflects the BDI model of agency in folk psychology~\cite{bdi}.
Also, as we later describe, modularizing the framework in this way allows for combining heuristic and data-driven components (e.g., neural networks). 

\subsection{Combining Differentiable and Heuristic Components}


Practically speaking, we need to make sure that the belief space $\mathcal{B}$ corresponds to something that can be understood as belief, that the intent space $\mathcal{I}$ corresponds to what can be understood as intent, etc. If we were to implement all three components with neural networks and train end-to-end, there is nothing that constrains the intermediate outputs to correspond to what we expect (e.g. no way to ensure the desire model output actually represents an ``intent"). 

We can mitigate this difficulty by imposing additional structure on the pipeline. The simplest way to do this is to replace one or more of the models with rule-based models, and/or to impose structural constraints on the input/output space of these models. In this paper we are primarily interested in inferring intent, so we choose to model the desire model with a neural network, while modeling the belief and action models with rule-based heuristics. 

Given a planning task we structure the belief state as a grid/graph of locations, and use a rule-based belief model to update this belief state given an observation. Additionally, we replace the action model with A* search, and structure the intents from the desire model as locations of subgoals.

In the setup described above, the belief and action models are rule-based, and the desire model is the sole trainable component. The rule-based belief model does not pose any issue with respect to the differentiability of the pipeline 
(one can think of it as simply preprocessing the input observation). However, we cannot optimize for the final output in any gradient-based way, as the output of the desire model is the input to the non-differentiable action model, which produces the final output. Unless we have ground-truth intents, training the desire model becomes difficult. 


\subsection{Inverse Action Model}
Given belief state $b \in \mathcal{B}$, observed action $a \in \mathcal{A}$, a set of intents $\mathcal{I}$, and non-differentiable action model $f: \mathcal{B} \times \mathcal{I} \rightarrow \mathcal{A}$, we want to learn a desire model $g$ to model a conditional distribution $p(z \mid b)$ such that
\begin{align*}
    \mathbb{E}[a \mid b] = \mathbb{E}_{z \sim g(b)}[f(b, z) \mid b]
\end{align*}
where $z \in \mathcal{I}$ is the intent. However, we may not have access to any samples from such a distribution (i.e. no ground truth $z \in \mathcal{I}$ for given $b, a$ pairs).

Alternatively, we can learn to model the distribution ${p(z \mid b, a)}$ with an ``inverse action model" $h$. This density of this distribution is proportional to
\begin{align}
    p(z \mid b, a) &\propto p(a, z \mid b) = p(a \mid b, z) \cdot p(z \mid b)
\end{align}
Because we have direct access to rule-based action model $f$ 
we can sample from $p(a \mid b, z)$, and thus given some kind of prior $p(z \mid b)$ we can sample from $p(z \mid b, a)$ to learn $h$.

Once we have learned an inverse action model $h$, then for each belief-action pair $(b, a)$ we can then simply use $h$ to sample intents from $p(z \mid b, a)$, and use these sampled intents to train the desire model $g$ in a supervised manner. 

\subsection{Training} 
Training is done in two stages: the first stage trains the inverse action model, the second stage trains the desire model. In each stage, once we gather the necessary data, we train using stochastic gradient descent (SGD).

\subsubsection{Training Inverse Action Model}
To train the inverse action model, we first collect belief states by sequentially running observations from human trajectories through the rule-based belief model and storing the resulting belief states at each timestep. These trajectories can be from human participants or potentially even from artificial agents trained to perform the task.

Now that we have collected these belief states, for each belief state $b$, we sample an intent $z$ given some prior $p(z \mid b)$, and create a set of $b, z$ pairs (for more implementation details on sampling and the prior, see the Appendix). Next, for each belief-intent pair, we generate an action $a = f(b, z)$, creating a dataset of belief-action-intent triples $(b, a, z)$.
Finally, we train the inverse action model on this dataset to predict intents given beliefs and actions; that is, using our generated dataset of $(b, a, z)$ tuples, given $b, a$, predict $z$.

Pseudocode for this training process is provided in Algorithm \ref{iamtrain}.

\subsubsection{Training Desire Model}
To train the desire model, we first collect belief states by running the observations from \textit{human} trajectories through the belief model. We also store the corresponding observed actions for each belief state.
We can then generate belief-intent pairs for the desire model by sampling intents from our inverse action model $z \sim h(b, a)$ for each belief-action pair. The target intents are formed by combining a) the probability distribution from the inverse action model over the previous belief state, and if available b) the next realized intent (from the future, in a post-hoc manner).
Finally we train the desire model on this data to predict intent given belief.

Pseudocode for this process is provided in Algorithm \ref{dmtrain}.

\begin{algorithm}
\SetAlgoLined
 \caption{Training inverse action model}
 \label{iamtrain}
 \SetKwInOut{input}{Input}
 \input{set of human trajectories $\mathcal{T} = \{\tau_1, \tau_2, \dots\}$, belief model $s$, action model $f$}
 $\mathcal{D}_{baz} \rightarrow \emptyset$ \\
 \For{$\tau \in \mathcal{T}$} {
    $o_1, \dots, o_n \rightarrow \tau$ \\
    $b_0 \rightarrow \textrm{Uniform}$ \\
    \For{$t = 1, \dots, n$} {
        $b_t \rightarrow s(b_{t-1}, o_t)$ \\
        \For{$i = 1, \dots, m$} {
            $z_t \sim p(z \mid b_t)$ \\
            $a = f(b_t, z_t)$ \\
            $\mathcal{D}_{baz} = \mathcal{D}_{baz} \cup \{(b_t, a, z_t)\}$
        }
    }
 }
 Initialize neural network parameters $\theta_h$  \\
 Use SGD to train $h(b, a \mid \theta_h)$ on dataset $\mathcal{D}_{baz}$
\end{algorithm}


\begin{algorithm}
\SetAlgoLined
 \caption{Training desire model}
 \label{dmtrain}
 \SetKwInOut{input}{Input}
 \input{set of human trajectories $\mathcal{T} = \{\tau_1, \tau_2, \dots\}$, belief model $s$, inverse action model $h$}
 $\mathcal{D}_{bz} \rightarrow \emptyset$ \\
 \For{$\tau \in \mathcal{T}$} {
    $(o_1, a_1), \dots, (o_n, a_n) \rightarrow \tau$ \\
    $b_0 \rightarrow \textrm{Uniform}$ \\
    \For{$t = 1, \dots, n$} {
        $b_t \rightarrow s(b_{t-1}, o_t)$ \\
        $z_t \sim h(b_t, a_t)$ \\
        $\mathcal{D}_{bz} = \mathcal{D}_{bz} \cup \{(b_t, z_t)\}$
    }
 }
 Initialize neural network parameters $\theta_g$  \\
 Use SGD to train $g(b \mid \theta_g)$ on dataset $\mathcal{D}_{bz}$
\end{algorithm}

\begin{figure}
    \centering
    \includegraphics[width=0.4\textwidth, height=70pt]{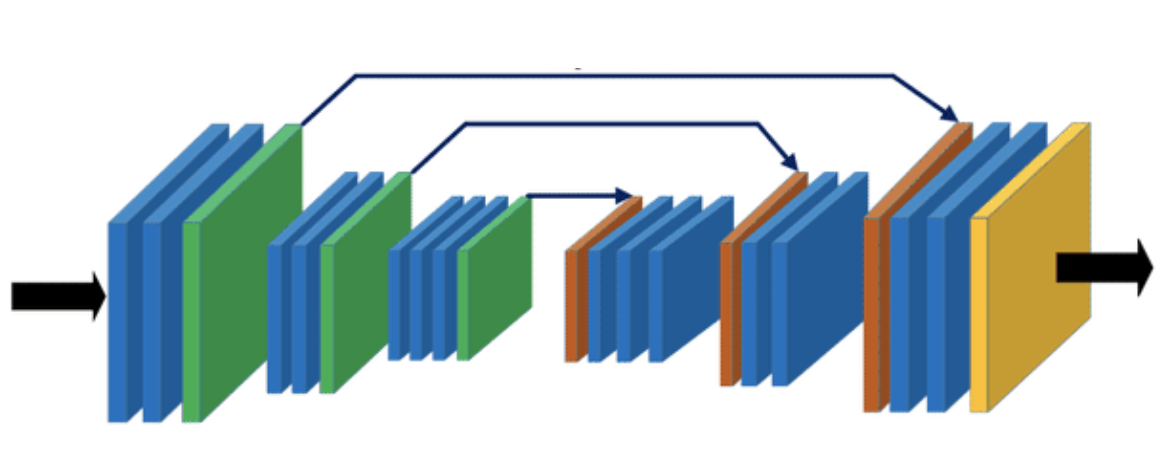}
    \caption{Encoder-decoder architecture used for the desire and inverse-action models in our experiments, inspired by U-Nets for image-segmentation \cite{ronneberger2015u}. Blue indicates convolutional layers, green indicates pooling, brown indicates up-sampling, and yellow is a final linear layer. A more detailed description can be found in the appendix.}
    \label{fig:desire}
\end{figure}

\section{CONCEPT WHITENING}

\textit{Concept whitening} (CW) is a mechanism introduced by Chen et al. \cite{conceptwhitening} for modifying neural network layers to increase interpretability. Broadly, it aims to enforce structure on the latent space by aligning the axes with predefined human-interpretable concepts. While this technique was developed for the purpose of image classification, here we adapt the idea in the context of intent inference with the desire model. By explicitly defining a set of concepts that can serve as ``explanations" for intent inferences, we can use concept whitening to allow for interpretability via identification of the most important concepts for any inference.

We also note that although we consider concept whitening in the context that can broadly be categorized as behavioral cloning, our approach to \textit{interpretable} agent-modeling is framework-agnostic and could potentially be applied to other reinforcement learning and imitation learning contexts.

\subsection{Technical Details}
Given latent representation $\mathbf{Z} \in \mathbb{R}^{n \times d}$, let $\mathbf{Z_C} \in \mathbb{R}^{n \times d}$ be the mean-centered latent representation. We can calculate the ZCA-whitening matrix $\mathbf{W} \in \mathbb{R}^{d \times d}$ as in \cite{iternorm}, and thus decorrelate and
standardize the data via whitening operation $\psi$:
\begin{equation}
    \psi(\mathbf{Z}) = \mathbf{WZ_C} = \mathbf{W}(\mathbf{Z} - \mu \mathbf{1}^\top)
\end{equation}
where $\mu = \frac{1}{n}\sum_{i=1}^n \mathbf{z}_i$ is the latent sample mean. 

Now say we are given concepts $c_1 \dots c_k$ that can be characterized by corresponding auxiliary datasets $\mathbf{X}_{c_1} \dots \mathbf{X}_{c_k}$, and assume we have an orthogonal matrix $\mathbf{Q} \in \mathbb{R}^{d \times d}$ such that the data from $\mathbf{X}_{c_j}$ has high activation on the $j$-th axis (i.e. column $\mathbf{q}_j$). Then the concept-whitened representation is given by:
\begin{equation}
    \hat{\mathbf{Z}} = \mathbf{Q^\top W Z_C}
\end{equation}

Training alternates between optimizing for the main objective (i.e.  the network's final output) and optimizing the orthogonal matrix $\mathbf{Q}$ for concept-alignment. To optimize $\mathbf{Q}$, we maximize the following objective:
\begin{equation}
    \max_{\mathbf{q_1} \dots \mathbf{q_k}} \sum_{j=1}^k \frac{1}{n_j} \sum_{\mathbf{x}_{c_j} \in \mathbf{X}_{c_j}} \mathbf{q}_j^\top \hat{\mathbf{z}}_{\mathbf{x}_{c_j}}
\end{equation}
where $\hat{\mathbf{z}}_{\mathbf{x}_{c_j}}$ denotes the concept-whitened latent representation in the model on  data sample from concept $c_j$. Orthogonality can be maintained when optimization is performed via gradient descent and curvilinear search on the Stiefel manifold \cite{wen2013feasible}. 
A more detailed description of concept whitening and the optimization algorithm can be found in \cite{conceptwhitening}.

\subsection{Concept Whitening for Intent Prediction}
We can modify this idea to the context of explanatory concepts for inferring intents. Specifically, we consider the desire model (Fig. \ref{fig:desire}) and insert a concept whitening layer (for more detail see the appendix and Fig. \ref{fig:desirecw}).

First we define a set of concepts $C = \{c_1, \dots, c_k\}$; these concepts should correspond to appropriate human-interpretable reasons or ``explanations" for intent prediction given the problem domain. We also must be able to identify a subset of timesteps from our trajectories where each concept applies, either directly from the trajectory data, or from external labels.

Recall that the desire model's inputs are belief states, which we can generate sequentially by passing the observations from each trajectory timestep through the belief model. Then for each concept $c_j$ we consider only the belief states from the timesteps where $c_j$ is known to apply, and aggregate them into auxiliary dataset $\mathbf{B}_{c_j}$.

Then training alternates between:
\begin{enumerate}
    \item Optimizing for intent prediction, given a belief state and a ground truth intent
    \item Concept-aligning the CW orthogonal matrix $\mathbf{Q}$ by maximizing the activation along axis $j$ for each auxiliary dataset $\mathbf{B}_{c_j}$
\end{enumerate}

Pseudocode for this process is provided in Algorithm \ref{desirecw}.

\begin{algorithm}
\SetAlgoLined
 \caption{Training desire model with concept whitening}
 \label{desirecw}
 $\mathcal{D} \rightarrow \emptyset$ \\
 \For{$\tau \in \mathcal{T}$} {
    $(o_1, a_1), \dots, (o_n, a_n) \rightarrow \tau$ \\
    $b_0 \rightarrow \textrm{Uniform}$ \\
    \For{$t = 1, \dots, n$} {
        $b_t \rightarrow \textrm{BM}(b_{t-1}, o_t)$ \\
        $z_t \sim \textrm{IAM}(b_t, a_t)$ \\
        $\mathcal{D} = \mathcal{D} \cup \{(b_t, z_t)\}$
    }
 }
 \For{$e = 1, ..., \mathit{num\_epochs}$} {
    Train DM on $\mathcal{D}$ with gradient descent \\
    \If{$e \mod 5 = 0$} {
        \For{$j = 1, ..., k$} {
            Maximize activation of $\mathbf{B}_{c_j}$ on the $j$-th column of $\mathbf{Q}$ \quad (see \cite{conceptwhitening})
        }
    }
 }
 
\end{algorithm}

\begin{figure*}
    \centering
    \includegraphics[width=0.64\linewidth]{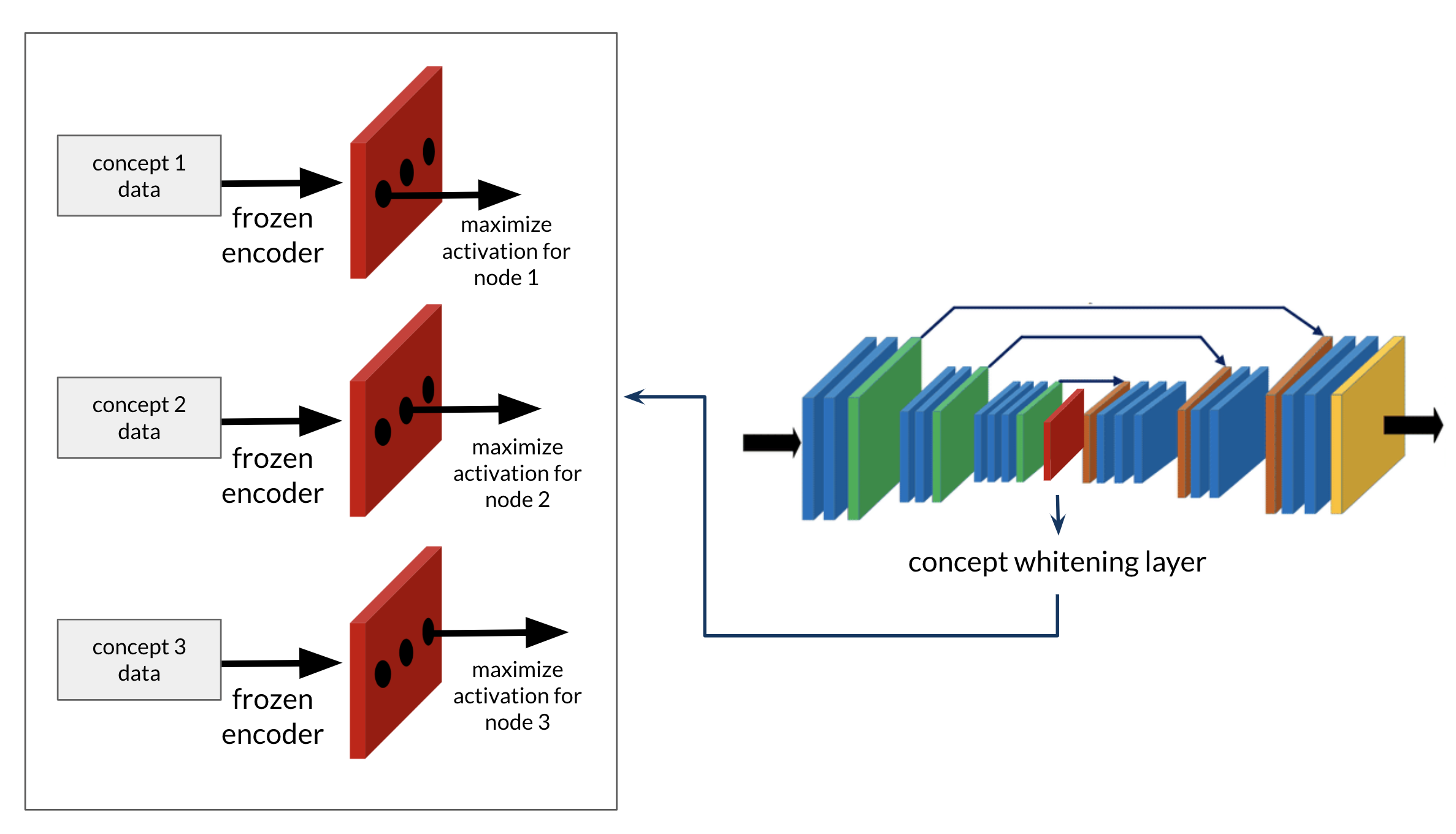}
    \caption{Desire model with concept whitening.}
    \label{fig:desirecw}
\end{figure*}

\section{EXPERIMENTS}

\subsection{Task Domain}
We consider a simulated search and rescue task in a Minecraft environment. The scenario simulates a damaged building after a disaster, with areas of the building layout perturbed with collapsed rubble, wall openings, and fires. There are 34 injured victims within the building who will die if left untreated. For convenience and simplicity, victims are represented as blocks. Out of these victims, 10 of these are critically injured and will expire after 5 minutes. These critical victims take 15 seconds to triage and are worth 30 points each. Other victims are considered ``non-critical", but will expire after 10 minutes. Non-critical victims take 7.5 seconds to triage and are worth 10 points each. The goal of the task is to earn as many points as possible within a 10 minute mission.

\subsection{Human Data For Training and Evaluation}
All experiments are performed using a set of 75 trajectories previously collected from human participants \cite{hsr}.
Prior to each mission, participants were given information on the task and the original building layout. However, the knowledge conditions of certain participants were manipulated by partially withholding information. Some participants were not informed of the cost-benefit tradeoffs (i.e. the knowledge that critical victims take 15 seconds to rescue and are worth 30 points and non-critical victims take 7.5 seconds to rescue and are worth 10 points). Knowledge was also manipulated via a beep signal that activated whenever the participant was near a room with a victim (1 beep for non-critical victim, 2 beeps for critical victim); certain participants were not told the meaning of the signal.

Participants were run in 3 knowledge conditions:
\begin{enumerate}
    \item No knowledge of critical-victim tradeoff, no knowledge of signal meaning
    \item Knowledge of critical-victim tradeoff but not of signal meaning
    \item Knowledge of both critical-victim tradeoff and signal meaning
\end{enumerate}

\subsection{Intent Prediction}
We represent intents as $(x, y)$ positions the participant intends to navigate towards. Specifically, we consider victims, doors, and room openings as locations-of-interest, which frames the intent prediction task as predicting either the \textit{next room} to be visited or the \textit{next victim} to be triaged. The predictions are accumulated at each timestep ($\sim$ 0.5 seconds per timestep) between visits of locations-of-interest, and then their mode is evaluated against the ground truth. We evaluate on a held-out test set of 20\% of participant trajectories.

\subsection{Concepts}

We defined a set of 10 concepts related to mission timer, knowledge condition, and field of view (see Table \ref{tab:concepts}). We consider 3 subsets:
\begin{itemize}
    \item Concept Set I is the full concept set
    \item Concept Set II omits the knowledge condition concepts
    \item Concept Set III omits both knowledge condition and mission timer concepts
\end{itemize}

\begin{table}[ht]
\caption{Concepts}
\label{tab:concepts}
\centering
\begin{tabular}{l|c}
    \textbf{Concept} & \textbf{Concept Sets} \\ \hline
    Mission timer between 0-3 minutes & I, II \\
    Mission timer between 3-5 minutes & I, II \\
    Mission timer between 5-8 minutes & I, II \\
    Mission timer $>$ 8 minutes & I, II \\
    Knowledge condition 1 (no triage, no signal) & I \\
    Knowledge condition 2 (triage, no signal) & I \\
    Knowledge condition 3 (triage, signal) & I \\
    Door / opening in field of view & I, II, III \\
    Non-critical victim in field of view & I, II, III \\ 
    Critical victim in field of view & I, II, III
\end{tabular}
\end{table}

The field of view and mission timer concepts were labeled directly from the data; the knowledge condition concepts are labeled with external knowledge of the condition for each participant trajectory.

\subsection{Results}

We compare the accuracy of ToM model intent predictions under 3 methods: training without Concept Whitening (CW), training from scratch with CW, and transfer learning by initializing a CW model with the weights of a pretrained non-CW model. Results are provided in Table \ref{ipacc}, where we see that introducing concept whitening for interpretability actually results in \textit{increased accuracy} of the model. 
The inductive bias produced by shaping the latent space to correspond to our selected concepts improves performance.

\begin{table}[ht]
\caption{Intent Prediction Performance 
}
\label{ipacc}
\begin{center}
\begin{tabular}{r|c}
\textbf{Training Method} & \textbf{Intent Prediction Accuracy} \\ \hline
Without CW & 73.0\% \\
\textbf{CW} & \textbf{84.0\%} \\
\textbf{CW + Transfer} & \textbf{84.1\%} \\
\end{tabular}
\end{center}
\end{table}

\subsection{Concept Ablation}

We also tested the effect of concept selection on performance (Table \ref{conceptablation}). In particular, we omitted the knowledge condition (KC) concepts and / or the mission timer concepts, tested concept-whitened ToM models both with and without transfer, and found noticeably diminished performance. 

Compared to the non-CW model, CW with reduced concept sets resulted in worse performance, and while transfer from the non-CW model somewhat mitigated this effect, we still see a significant drop from the performance with full concept set. This demonstrates the importance of good concept selection for the resulting performance of concept-whitened ToM model.

\begin{table}[ht]
\caption{Varying Concept Sets}
\label{conceptablation}
\begin{center}
\begin{tabular}{r|ccc}
\textbf{Training Method} & \textbf{Concept Set} & \textbf{Acc.} \\ \hline
Without CW & N/A & 73.0\% \\
CW & III & 41.2\% \\
CW & II & 69.2\% \\
\textbf{CW} & \textbf{I} & \textbf{84.0\%} \\
CW + Transfer & III & 54.9\% \\
CW + Transfer & II & 77.9\% \\
\textbf{CW + Transfer} & \textbf{I} & \textbf{84.1\%} \\
\end{tabular}
\end{center}
\end{table}

\section{INTERPRETABILITY ANALYSIS}

Generally, using a deep model for a task such as intent inference makes interpreting model output difficult due to the ``black-box" nature of neural networks. Adding the concept whitening module provides us with a direct mechanism that allows us to interpret the model output in terms of our chosen concepts. For each observation and corresponding intent prediction made by our ToM model, we can calculate the importance of each concept.

For example, if our ToM model observes that a player hears a beep and predicts the player intends 
to bypass a particular door, we can examine the relative importance of concepts such as ``critical victim in field of view" or ``no knowledge of the beep signal's meaning" or ``over 8 minutes elapsed in mission timer" for our model's prediction. If for instance the most important concept is ``no knowledge of beep signal's meaning", an appropriate intervention could be to inform the player of the signal's meaning.

\subsection{Qualitative Analysis}
We can estimate the concept importance for each prediction via the activation for each column of the CW orthogonal matrix $\mathbf{Q}$, given by:
\begin{equation}
    a_j = \mathbf{q}_j^\top \hat{\mathbf{z}}_{b}
\end{equation}
where $\hat{\mathbf{z}}_{\mathbf{x}}$ is the concept-whitened latent representation for belief state $b$.

\begin{figure}[htp]
    \centering
    \begin{subfigure}{\linewidth}
        \includegraphics[width=\linewidth]{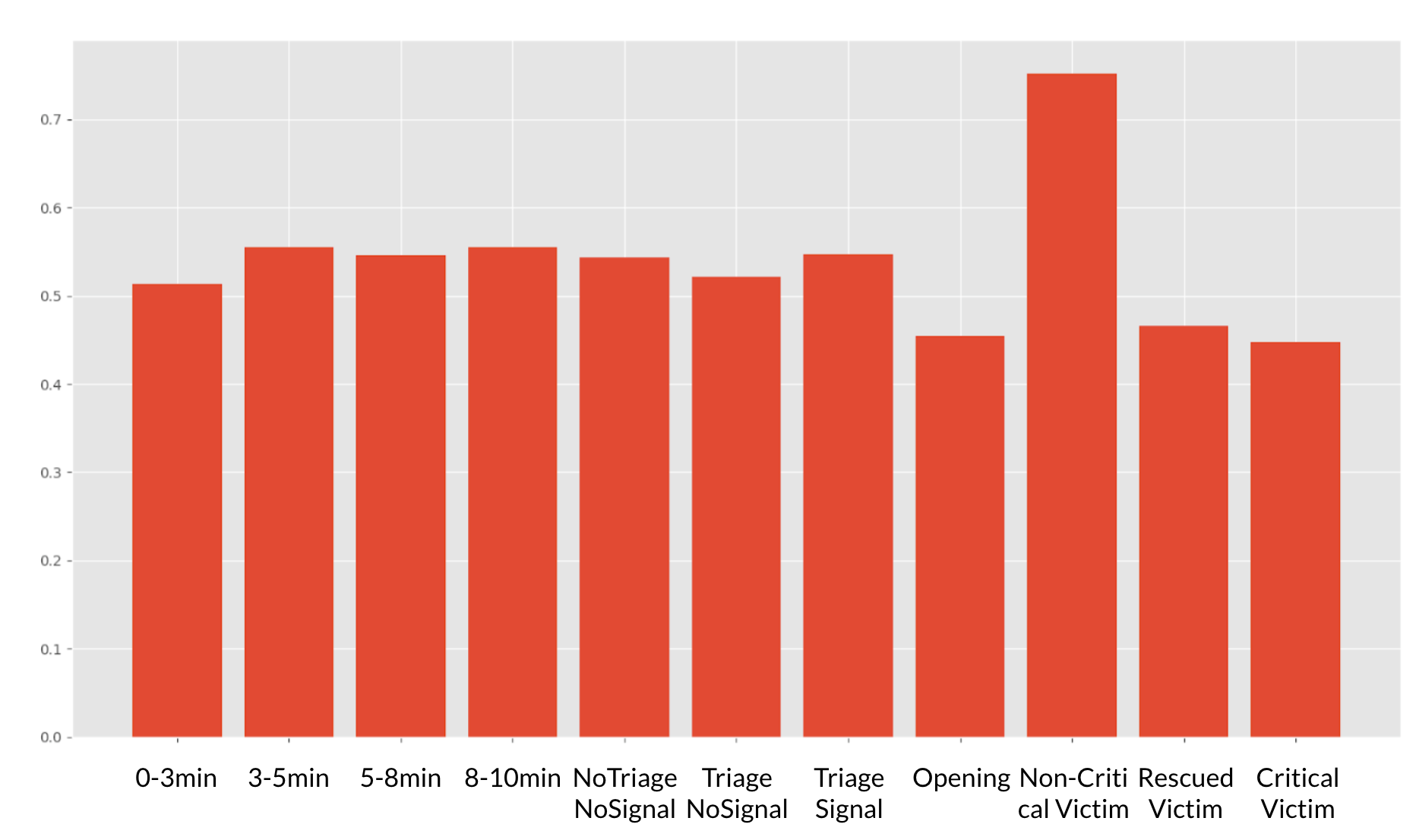}
        \caption{Mean concept activation for intent prediction of \textbf{non-critical victim}. We can see that the presence of a non-critical victim in the field-of-view is the most activated concept.}
        \label{fig:green_activation}
    \end{subfigure}
    \vspace{20pt}
    
    \begin{subfigure}{\linewidth}
        \includegraphics[width=\linewidth]{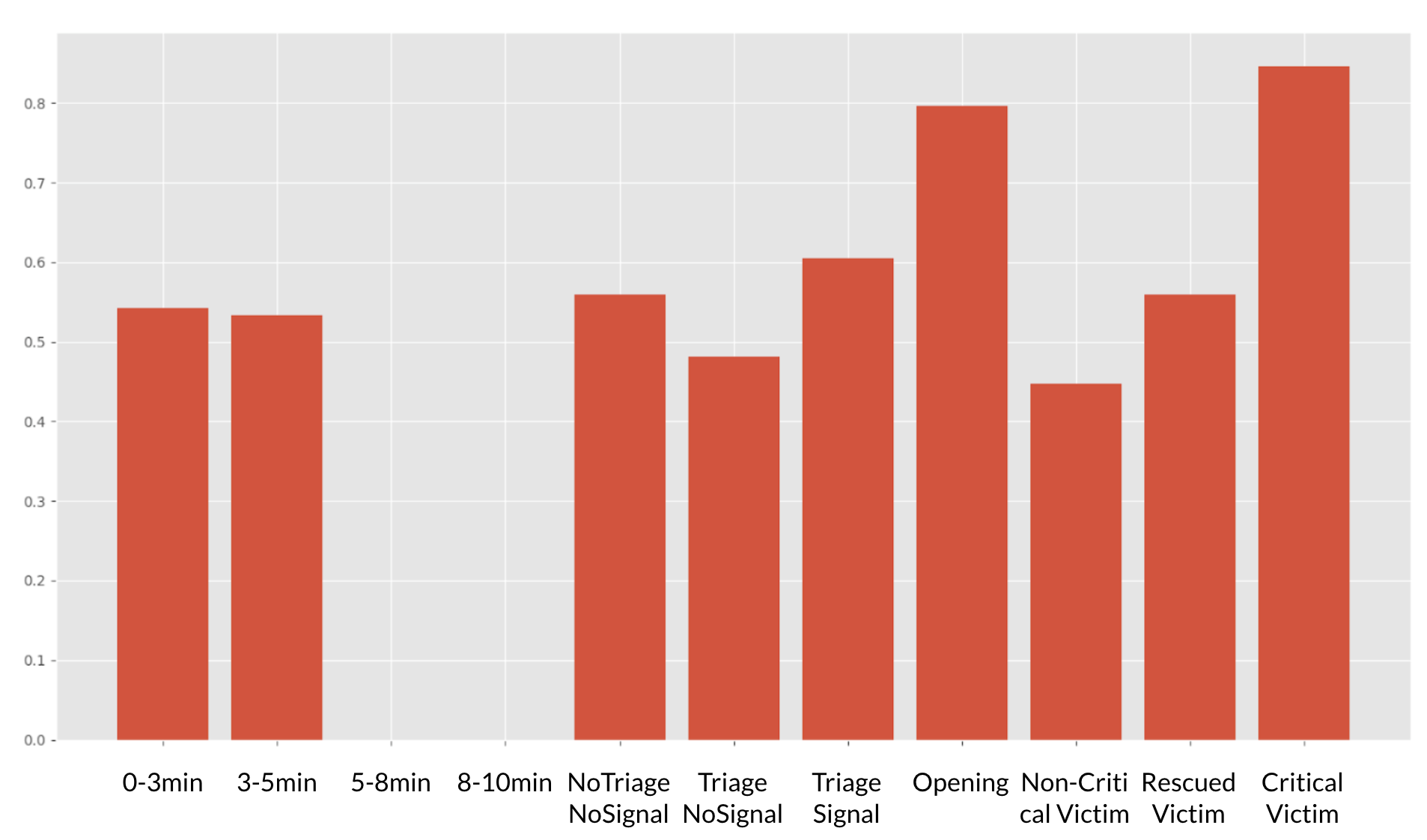}
        \caption{Mean concept activation for intent prediction of \textbf{critical victim}. Here we see zero activation for mission timer above 5 minutes (which corresponds with critical victims expiring). We also see that the presence of a critical victim or room opening in field-of-view is a common reason for predicting intent to triage a critical victim.}
        \label{fig:yellow_activation}
    \end{subfigure}
    \vspace{20pt}
    
    \begin{subfigure}{\linewidth}
        \includegraphics[width=\linewidth]{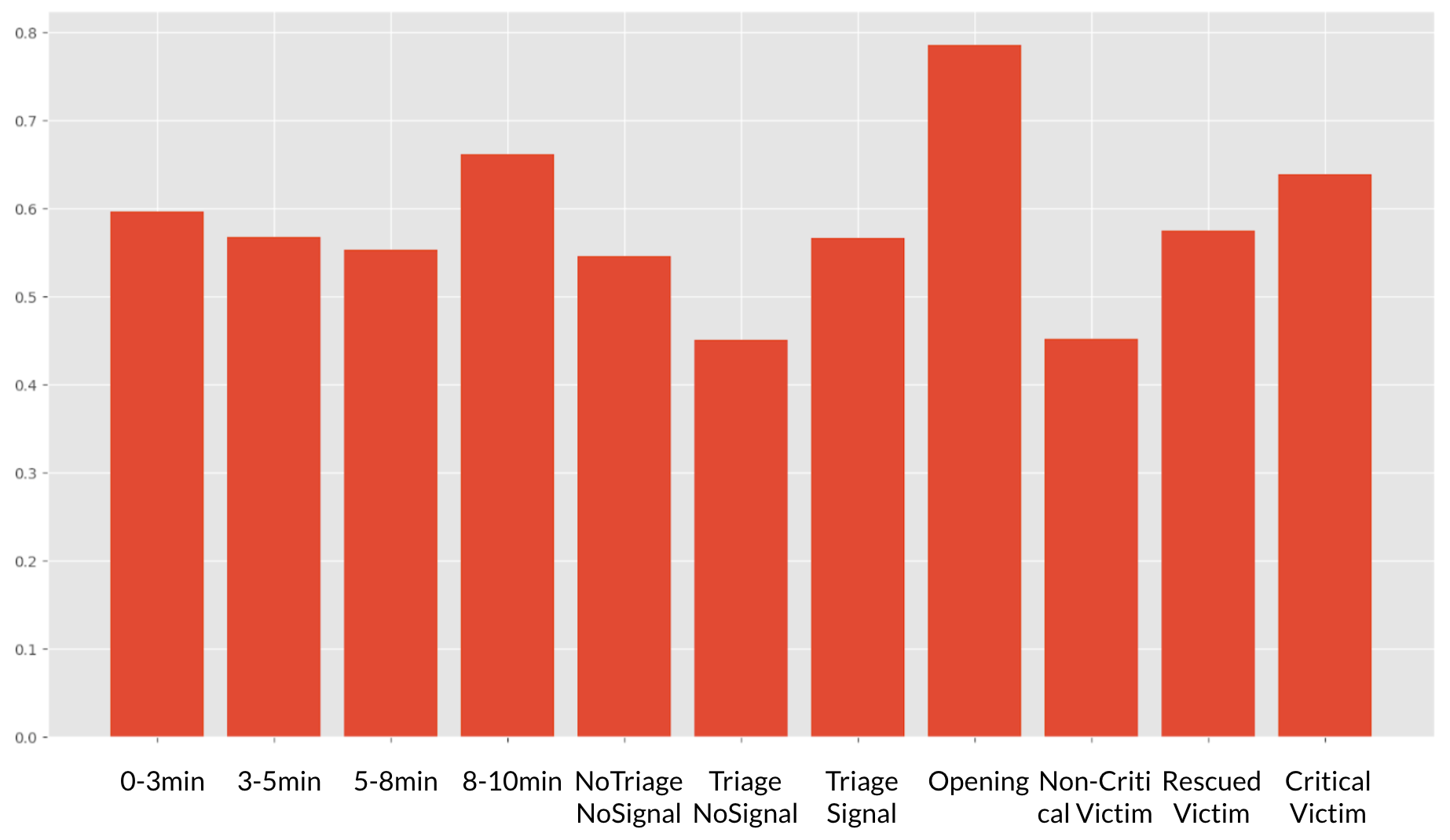}
        \caption{Mean concept activation for intent prediction of \textbf{opening}. The presence of an opening in the field of view is the most highly activated concept. We also see that compared to the other mission timer concepts, the last 2 minutes sees the timer become a more important reason for predicting intent to go towards a door or opening.}
        \label{fig:opening_activation}
    \end{subfigure}
    \caption{Mean concept activations for different intent prediction types.}
    \label{fig:activation}
\end{figure}

We can examine the activation vectors $\mathbf{a} = [a_1 \dots a_k]$ for different types of intent predictions by the learned model (CW + transfer, full concept set). The mean normalized activations for non-critical victims, critical victims, and doors / openings are visualized in Fig. \ref{fig:activation}.

These largely line up with intuition; unsurprisingly, the presence of an intent-relevant entity in the field of view is an important concept for the model's prediction of said intent. We also see variability in the importance of different mission time intervals and of knowledge condition for different intent predictions. 


\subsection{Quantitative Analysis}

We also attempt to quantitatively assess  how well the concept activation vectors characterize the ToM model's intent inferences. Intuitively, we should be able to deduce information about what the model’s prediction will be with high accuracy, using only concept activation vectors as input.

This can be framed as a classification problem where, given an activation vector, we predict the type of the corresponding predicted intent as one of: non-critical victim, critical victim, or opening (i.e. door or hole in the wall between rooms). Rather than use a complex model, 
we instead learn simpler models -- a decision tree and an SVM -- and use the accuracy as a proxy for the quality of our concept activations. As shown in Table \ref{decisiontreesvm}, we achieve relatively high accuracies with these simple models.

\begin{table}[ht]
\caption{Classifiying Activations as Intents}
\label{decisiontreesvm}
\begin{center}
\begin{tabular}{r|c}
\textbf{Model} & \textbf{Accuracy} \\ \hline
Decision Tree & 93.0\% \\
SVM & 92.0\% \\
\end{tabular}
\end{center}
\end{table}

\section{CONCLUSION}

We have presented a modular ToM approach for reasoning about humans  that can allow for both neural and heuristic components. Our approach explicitly incorporates interpretability while still maintaining the performance and scalability advantages of neural approaches. We move beyond simple toy environments and apply our framework to a more complex setting.  Our experimental results demonstrate that enforcing interpretability can also increase predictive accuracy.

The natural extension of this work is in exploring the benefits of ToM and interpretability in assistance and interventions. A particularly interesting direction would be to explore \textit{counterfactuals}. This means examining how intent inference changes given a change in concept activation, and then finding the closest belief state that could result in the given change. Approaching this through interpretable concept activations rather than in the belief space could facilitate \textit{interventions} to warn about or correct human errors when working in Human + AI teams.

Another potential avenue of future work is in artificial agents explaining their own decisions and interventions in terms of human-interpretable concepts. We hope that this work can serve as a starting point for social intelligence in AI systems that are designed to interact \textit{with humans}.

\section*{APPENDIX}

\subsection{Implementation Details}
We consider Minecraft environments that can effectively be 
projected onto 2D grid representations (e.g., building interiors). The specification for each of the framework components is given below.

\paragraph*{Observations} Observations are represented $X \times Y$ grids (where $X$ and $Y$ are dimensions of the environment), and each $(x, y)$ coordinate contains one of $K$ different block types.

\paragraph*{Belief States}
Each \textit{belief state} $b$ is represented by a $X \times Y \times K$ grid, where the value at $(x, y, k)$ represents the probability of the block at position $(x, y)$ having block type $k$.

\paragraph*{Belief Model}
We use a rule-based \textit{belief model} that aggregates observations into our belief state with probability 1, and decay probabilities over time to a uniform distribution over block types by $b \rightarrow \frac{b + \epsilon}{1 + K\epsilon}$ after each timestep, where $b$ is a belief state grid, $K$ is the number of block types, and $\epsilon$ is a forgetfulness hyperparameter we set to $0.01$.

\paragraph*{Intents}
We represent each \textit{intent} as an $(x, y)$ position the player intends to navigate towards.

\paragraph*{Intent Prior}
When generating data to train the inverse action model, for each belief state $b$, we sample an intent $(x, y)$ given some prior $p(x, y | b)$, and create a set of $b, (x, y)$ pairs. We specifically use the prior $p(x, y | b) = \frac{1}{d_b(x, y)}$ if we belief a victim or door is at position $(x, y)$, and 0 otherwise, where $d_b(x, y)$ is the L1 distance of point $(x, y)$ from the player's position. 

\paragraph*{Action Model}
We use A* search as our \textit{action model}, $\textrm{A}^*(b, (x, y)) = a$, where $b$ is a belief state, $(x, y)$ represents the intent, and $a$ is an action from discrete action set of: \texttt{left\_turn}, \texttt{right\_turn}, \texttt{toggle\_door}, \texttt{toggle\_lever}, \texttt{triage}, or \texttt{None}.

\subsection{Neural Architectures}

\paragraph*{Inverse Action Model} The inverse action model takes as input a belief state $b$ and an action $a$. It outputs an $X \times Y$ grid of log-probabilities for the intent at each grid cell. It is designed as an encoder-decoder model, inspired by image-segmentation approaches, and uses the following architecture (Fig \ref{fig:desire}):
\begin{itemize}
    \item 3 encoder blocks each consisting of a convolutional layer, followed by max-pooling, ReLU activation and batch norm
    \item A bottleneck layer, concatenating the downsampled input and action before passing through a linear layer
    \item 3 decoder blocks each consisting of:
    \begin{itemize}
        \item a deconvolutional upsampling layer
        \item a residual connection with the output of the corresponding encoder block
        \item a convolutional layer, followed by ReLU activation and batch norm
    \end{itemize}
\end{itemize}

\paragraph*{Desire Model} The desire model takes a belief state $b$ as input.
It outputs an $X \times Y$ grid of log-probabilities for the intent at each grid cell. Its architecture (Fig \ref{fig:desire}) is identical to that of the inverse action model, except without concatenating the action in the bottleneck layer. When training concept whitening, we replace batch normalization after the bottleneck layer with a concept whitening layer.



\bibliography{citations}
\bibliographystyle{IEEEtran}

\end{document}